# ENHANCED PROTOTYPICAL PART NETWORK (EPPNET) FOR EXPLAINABLE IMAGE CLASSIFICATION VIA PROTOTYPES

Bhushan Atote and Victor Sanchez
Department of Computer Science, The University of Warwick, UK
E-mail: {bhushan.atote, v.f.sanchez-silva}
@warwick.ac.uk

**Abstract**

Explainable Artificial Intelligence (xAI) has the potential to enhance the transparency and trust of AI-based systems. Although accurate predictions can be made using Deep Neural Networks (DNNs), the process used to arrive at such predictions is usually hard to explain. In terms of perceptibly human-friendly representations, such as word phrases in text or super-pixels in images, prototype-based explanations can justify a model's decision. In this work, we introduce a DNN architecture for image classification, the Enhanced Prototypical Part Network (EPPNet), which achieves strong performance while discovering relevant prototypes that can be used to explain the classification results. This is achieved by introducing a novel cluster loss that helps to discover more relevant human-understandable prototypes. We also introduce a faithfulness score to evaluate the explainability of the results based on the discovered prototypes. Our score not only accounts for the relevance of the learned prototypes but also the performance of a model. Our evaluations on the CUB-2002011 dataset show that the EPPNet outperforms state-of-theart xAI-based methods, in terms of both classification accuracy and explainability.

## 1. INTRODUCTION

Despite their remarkable success in several domains [1, 2, 3, 4, 5], the majority of DNNs operate as "black boxes," making their decision-making processes obscure [6]. This opacity may result in training models that generate biased results toward specific data types. Hence, it is important for AI to be explainable not only to allow for a clear understanding of a model's results but also to ensure accountability in compliance with existing laws and regulations. [7, 8, 9, 10].

The Defense Advanced Research Projects Agency (DARPA) highlights the main objective of xAI as developing models that are explainable and can attain strong performance [11]. Such models should ensure that users can effectively interact with and trust AI technologies. Therefore, they should provide explanations that are clear to all users, regardless of their technical expertise. Based on such requirements, xAI-based methods can be categorized into those that provide active (intrinsic) and passive (post-hoc) explanations [12, 13, 14, 15]. Active methods involve modifying a model's architecture or changing the training strategy to enhance explainability. On the other hand, passive methods do not modify a model's architecture or its training process. Instead, they attempt to derive logical rules or identify comprehensible patterns based on a trained model.

Among the active xAI-based methods, those that merge prototype-driven explanations with the learning capacity of neural networks have been shown to provide appropriate explanations that can be understood by non-expert users. This is usually accomplished by measuring the similarity of the results to representative prototypes learned during training [16]. In the context of imaging data, such prototypes are usually relevant regions of the training images. For example, the PrototypeDNN merges an autoencoder with a prototype layer, offering explanations for predictions in the form of important prototypes identified during training.[16]. Alvarez et al.'s SENN introduces a shift from simple linear classifiers to transparent, complex models, meeting explainability standards of explicitness, faithfulness, and stability [17]. ProtoPNet (PPNet) and ProtoPShare provide human-like reasoning in image classification and enhance prototype consistency, while Nauta et al. propose amplifying explainability by pairing visual prototypes with textual data [18, 19, 20]. ProtoPool and Deformable ProtoPNet, on the other hand, introduce innovations in prototype-based models, focusing on vital visual aspects and spatial flexibility [21, 22].

Based on the success of prototype-based explanations, we introduce a new active xAI-based method: the Enhanced Prototypical Part Network (EPPNet), a DNN architecture for image classification that explains its results by learning relevant prototypes during training. The EPPNet, which is based on the PPNet, provides improved explanations for each prediction while improving the overall classification performance. This is achieved by introducing a novel mean-cluster loss that leverages several prototypes learned for each class. We also introduce a new explainability metric: the faithfulness score. This metric provides insights into the relevance of prototypes to explaining the classification results, on a per-class basis, while accounting for the performance of the model. Our experiments on the CUB-200-2011 dataset indicate that 1) compared to the PPNet, the prototypes learned by the EPPNet are not only more relevant but also maintain a robust correlation with the training images available for each class, and 2) the EPPNet can outperform other active xAI-based methods in terms of classification results.

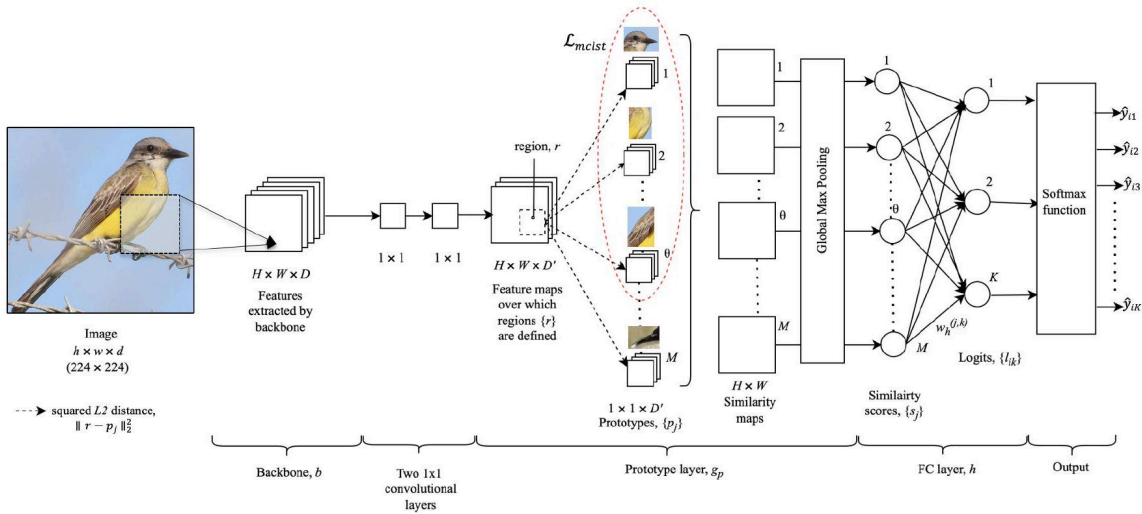

Fig. 1. The Enhanced Prototypical Part Network (EPPNet), which uses a novel mean cluster loss, $\mathcal{L}_{\text{mclst}}$, relies on $\theta > 1$ squared $L2$ distances, each computed between any region $r$ defined over the extracted feature maps and a prototype $p_j$.

## 2. PROPOSED ARCHITECTURE

The architecture of the EPPNet is illustrated in Fig. 1. It is based on the architecture of the PPNet [18], with important modifications to the prototype layer. Specifically, it introduces a mean-cluster loss, which provides superior classification and explainability performance.

The proposed EPPNet comprises three key components: a pre-trained convolutional backbone network, $b$, which extracts feature maps from the image $x_i$, a prototype layer, $g_p$, and a fully connected (FC) layer, $h$, leading to the output logits, which are then mapped to probabilities by a Softmax function to determine the class of $x_i$. The prototype layer learns $M$ prototypes for $K$ classes. Each prototype represents a significant activation pattern within the extracted feature maps and corresponds to an image region in the original pixel domain. Within this context, the $j^{th}$ prototype, denoted by $p_j$, can be seen as the latent representation of a prototypical part of an image. The prototype layer, $g_p$, calculates the squared $L2$ distance between each $p_j$ and each possible region $r$ defined over the extracted feature maps, where $r$ has the same dimensions as $p_j$. The resulting distance values are organized into similarity maps after inverting them (see Fig. 1). The activation maps are then reduced to a similarity score by a global max pooling operation. Each similarity score, denoted by $s_j$, quantifies how strongly prototype $j$ is present in some region of the image $x_i$ [18]. The highest similarity score $s_j$ is attained for the smallest square $L2$ distance. The $M$ similarity scores, $\{s_j\}$, are then used to compute the logit of image $x_i$ associated with class $k \in K$:

$$l_{ik} = \sum_{j=1}^{M} s_j \cdot w_h^{(j,k)} \qquad (1)$$

where $w_h^{(j,k)}$ is the weight of the FC layer associated with the $j^{th}$ prototype and class $k$. Finally, these logits are normalized using a Softmax function to provide the probabilities of an image belonging to the existing classes.

Like the PPNet, once trained, the EPPNet predicts the class of an input image by evaluating how closely it resembles the prototypes learned for each class. Such a similarity is quantified by the similarity scores. Hence, the similarity scores highlight the prototype's role in the prediction and can be used to understand and explain the classification results.

### 2.1. Mean-cluster loss

The total loss used to train the proposed EPPNet is:

$$\mathcal{L}_{\text{total}} = \mathcal{L}_{ce} + \lambda_1 \mathcal{L}_{mclst} + \lambda_2 \mathcal{L}_{sep} \qquad (2)$$

where $\mathcal{L}_{ce}$ is a cross-entropy loss, $\mathcal{L}_{mclst}$ is the proposed mean cluster loss, $\mathcal{L}_{sep}$ is a separation cost, and $\{\lambda_1, \lambda_2\}$
are hyper-parameters used to weigh the contributions of the losses. These three losses are computed as follows:

$$\mathcal{L}_{ce} = -\frac{1}{n}\sum_{i=1}^{n}\left(\sum_{k=1}^{K} y_{ik}\log(\hat{y}_{ik})\right) \qquad (3)$$

$$\mathcal{L}_{mclst} = \frac{1}{n}\sum_{i=1}^{n}\left(\frac{1}{\theta}\sum_{\omega=1}^{\theta}\min_{\substack{r_\omega \in \phi(x_i);\\ j:p_j \in P_{y_{ik}}}} \|r_\omega - p_j\|_2^2\right) \qquad (4)$$

$$\mathcal{L}_{sep} = -\frac{1}{n}\sum_{i=1}^{n}\min_{j:p_j \notin P_{y_{ik}}}\min_{r\in\phi(x_i)} \|r - p_j\|_2^2 \qquad (5)$$

where $n$ is the number of training images, $y_{ik} \in \{0,1\}$ is the ground truth value of image $x_i$ for class $k$, i.e., $y_{ik}=1$ denotes that image $x_i$ belongs to class $k$, $\hat{y}_{ik} \in [0,1]$ is the predicted probability for the $k^{th}$-class, $P_{y_{ik}}$ is the set of prototypes of the same class as image $x_i$, $\phi(x_i) = \{r\}$ is the set of regions defined over the feature maps extracted from $x_i$, and $\mathcal{L}_{mclst}$ is our mean-cluster loss function that returns the mean of the $\theta$ smallest squared distances between any region in set $\phi(x_i)$ and any prototype in set $P_{y_{ik}}$. For example, for $\theta = 3$, our mean-cluster loss returns $\frac{1}{3}(d_1 + d_2 + d_3)$, where $d_1$ is computed as follows:

$$d_1 = \min_{j:p_j \in P_{y_{ik}}}\min_{r\in\phi(x_i)} \|r - p_j\|_2^2 \qquad (6)$$

while $d_2$ and $d_3$ are computed the same way as $d_1$ but excluding from $\phi(x_i)$, respectively, the region $r_1$ used to compute $d_1$, and the regions $\{r_1, r_2\}$ used to compute $d_1$ and $d_2$. In other words, in this example, $d_1$ is the smallest squared $L2$ distance, while $d_2$ and $d_3$ are the second and third smallest squared $L2$ distances, respectively.

The losses $\mathcal{L}_{ce}$ and $\mathcal{L}_{\text{sep}}$, are used, respectively, to assign image $x_i$ to the correct class and to prevent the regions defined over the features maps of $x_i$ from being associated with prototypes of incorrect classes [18]. These two losses are also used by PPNet. The mean-cluster loss, $\mathcal{L}_{\text{mclst}}$, on the other hand, differs from the cluster cost, $\mathcal{L}_{\text{clst}}$, used by the PPNet, which is as follows:

$$\mathcal{L}_{\text{clst}} = \frac{1}{n}\sum_{i=1}^{n}\min_{j:p_j \in P_{y_{ik}}}\min_{r\in\phi(x_i)} \|r - p_j\|_2^2 \qquad (7)$$

Note that $\mathcal{L}_{\text{clst}}$ encourages the PPNet to assign any region $r$ defined over the features maps extracted from a training image to one prototype of the same class as the image. This assignment is also based on the squared $L2$ distance between $r$ and $p_j \in P_{y_{ik}}$ [18]. However, such a strategy may overlook the fact that multiple prototypes can be associated with multiple regions of a training image, which has the potential to improve explainability and classification accuracy. For example, let use consider the set of squared distances, $\{d_1 = 0.010, d_2 = 0.020, d_3 = 0.040, d_4 = 0.018, d_5 = 0.030\}$, loss $\mathcal{L}_{\text{clst}}$ is then likely to form a cluster with the region and prototype used to compute the minimum distance $d_1$. On the other hand, our loss $\mathcal{L}_{mclst}$ in Eq. 4 is likely to form a cluster of $\theta$ regions and corresponding $\theta$ prototypes. In this example, if $\theta = 3$, the formed cluster uses the regions and prototypes that are used to compute distances $d_1, d_2$ and $d_4$, which provides more information about the similarities between an image and the prototypes. This strategy allows discovering prototypes that are richer in information because during training, several prototypes can be updated based on each training image. This idea is illustrated in Fig. 2.

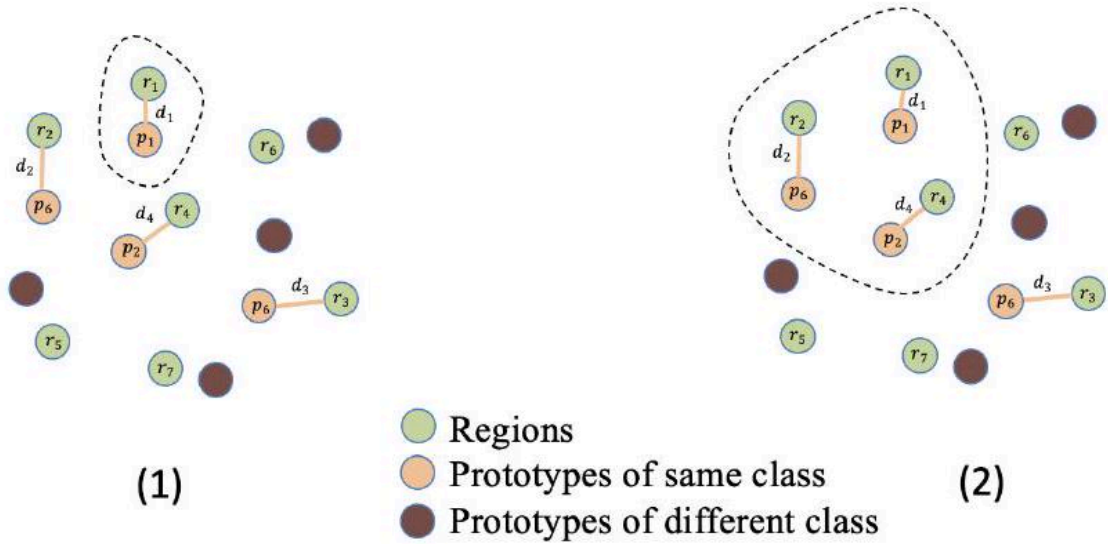

Fig. 2. Cluster formation in (1) the PPNet and (2) the EPPNet when $\theta = 3$. Note that the PPNet only uses $r_1$ and $p_1$ in this example, as this pair results in the minimum squared $L2$ distance between any $\{r\}$ and any $\{p_j\}$ of the same class. The EPPNet, on the other hand, also uses the second and third minimum distances; i.e., $d_2$ and $d_4$ in this example.

Let us recall that the prototype layer, $g_p$, computes the squared $L2$ distance between $\{r\}$ and $\{p_j\}$. Fig. 3 shows a set of such squared distances plotted in dotted lines as the training process progresses. Let us now consider the red curve in this figure, which represents the result of a function $\mu$ that computes the minimum value of the set of squared $L2$ distances. Let us also consider the blue curve, which represents the result of a function $\nu$ that computes the minimum value of the mean of the same set of squared $L2$ distances. One can observe that the functionality of the loss $\mathcal{L}_{clst}$ can be represented by the red curve, i.e., the $\mu$ function, while the functionality of our proposed mean-cluster loss $\mathcal{L}_{mclst}$ can be presented by the blue curve, i.e., the $\nu$ function. Note that several local minima are evident in the red curve, which makes finding the global minimum challenging. Conversely, the blue curve exhibits a smoother profile with less sharp local minima, hence making it easier to find the global minimum. Consequently, by using $\theta > 1$ squared $L2$ distances in our proposed loss $\mathcal{L}_{mclst}$, a close approximation to the global minimum can be found more easily, which helps to train the EPPNet and learn more relevant prototypes.

## 2.2. Faithfulness score

Our proposed faithfulness score aims at answering the question: [17]: Are relevant features (i.e., prototypes) genuinely

5 / 13

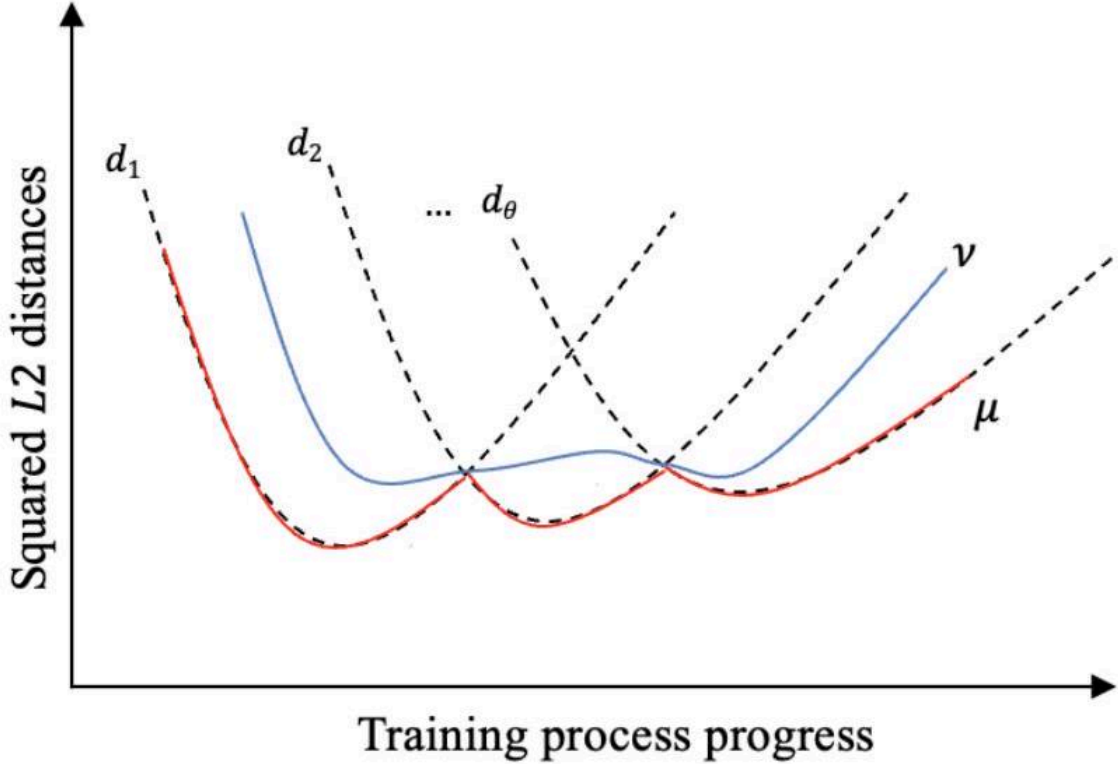

Fig. 3. Example curves representing a set of squared distance distances $\{d_1, d_2, \cdots, d_\theta\}$ as the training process progresses (i.e., number of epochs). The red curve, $\mu$, is the result of finding the minimum value of this set, while the blue curve, $\nu$, is the result of finding the minimum value of the mean of this set.

relevant?. To this end, our score is computed based on the logits, which are derived from the class connections and similarity scores, as given by Eq. 1. Note that using the logits to define our score instead of the Softmax probabilities affords two main advantages: 1) Once the model is trained and the prototypes are learned, the logits provide information about how useful the learned prototypes are to assign an image a specific class label. And 2), the value of a logit is expected to be high if the corresponding similarities scores are high and the weights $w_h^{(j,k)}$ learned for the $FC$ layer are also high. Hence, the value of a logit implicitly carries information about the accuracy of the model. In other words, the learned prototypes are strongly related to the training images and strongly linked to a specific class, $k$, if the corresponding logit value is high. A score based on the logits can then help to determine if the prototypes learned during training are relevant to classifying the images.

Based on the previous observations, to estimate the faithfulness score of class $k$, denoted by $t_k$, we use all the test images of class $k$:

$$t_k = \frac{1}{z_k}(f(\mathbf{l}_1) \cdot \max(\mathbf{l}_1) + f(\mathbf{l}_2) \cdot \max(\mathbf{l}_2) + \ldots \\ \ldots + f(\mathbf{l}_{zk-1}) \cdot \max(\mathbf{l}_{zk-1}) + f(\mathbf{l}_{z_k}) \cdot \max(\mathbf{l}_{z_k})) \qquad (8)$$

where $z_k$ denotes the number of test images of class $k$, $\mathbf{l}_i = \{l_{i1}, l_{i2}, \cdots, l_{iK}\}$ is the set of $K$ logits for the $i^{th}$ test image; i.e., one logit per class, and $f(\cdot)$ is function that returns 1 if test image $i$ is classified correctly and -1, otherwise.

Note our faithfulness score in Eq. 8 is normalized by the total number of test images in the class, i.e., by $z_k$. Such normalization allows for the computation of a score per class that accounts for cases when some classes may have more test images than others. Also, note that our score uses the highest logit values in all $\{l_i\}$. By using this set of $z_k$ highest logit values, our score accounts for the relevance of the learned prototypes used to classify the test images into class $k$. In other words, it uses the strongest association and similarity between the highest activated regions of the feature maps extracted from a test image and the learned prototypes for each class. The score is a summation of such highest logit values where the sign to be used in the summation is given by function $f(\cdot)$. Consequently, a correctly classified test image is assigned a positive value in the sum while an incorrectly classified test image is assigned a negative value. Such a strategy accounts for the performance of the model on the test images. A large and positive faithfulness score $t_k$ then indicates a robust linkage of prototypes to class $k$, emphasizing their high relevance to that class and the strong performance of the model in that class. Conversely, a low and negative faithfulness score $t_k$ indicates poor performance and low relevance of the learned prototypes for that class.

# 3. EXPERIMENTS

Dataset: We use the CUB-200-2011 [23] dataset for evaluation, with all RGB images adjusted to a size of $224 \times 224$. This dataset is an extensive collection of bird images with 200 classes. It is important to clarify that the CUB-200-2011 dataset includes part location labels manually identified for the images. Let us recall that the primary goal of active xAIbased methods, like the one proposed in this work, is to generate explanations, i.e., prototypes, without relying on any prior knowledge such as these part location labels. In other words, the aim is to discover the prototypes through the learning process. Hence, these part location labels are not useful in our evaluations.

Implementation details: We use VGG-19, ResNet-34, ResNet-152, DenseNet-121, and DenseNet-161 as the backbone for the proposed EPPNet and other evaluated methods. The backbone is followed by two $1 \times 1$ convolutional layers. Except for the final $1 \times 1$ convolutional layer, which uses a Sigmoid activation function, all other previous convolutional layers use the ReLU activation function. The EEPNet and the PPNet are trained to identify $M = 2000$ distinct prototypes, 10 per class for $K = 200$ classes, where each prototype has dimensions $1 \times 1 \times D'$ in the latent space, where the depth $D'$ matches that of the output of the convolutional layers. Let us recall that the prototypes capture representative activation patterns within the extracted feature maps, which in turn reflect prototypical regions within the pixel domain. We run the experiments for 100 epochs. In the total loss of the EPPNet, the mean cluster loss, $\mathcal{L}_{\text{mclst}}$, and the separation cost, $\mathcal{L}_{\text{sep}}$, are weighted by $\lambda_1 = 0.8$ and $\lambda_2 = -0.8$, respectively.

Training algorithm: Like the PPNet, the EPPNet is trained is three stages. The first stage updates all weights except for those of the FC layer. The second stage adjusts prototypes $\{p_j\}$ by associating them with the corresponding regions $\{r\}$ that minimize the squared $L2$ distance. The third stage updates the weights of the FC layer, freezing all other weights until convergence is achieved.

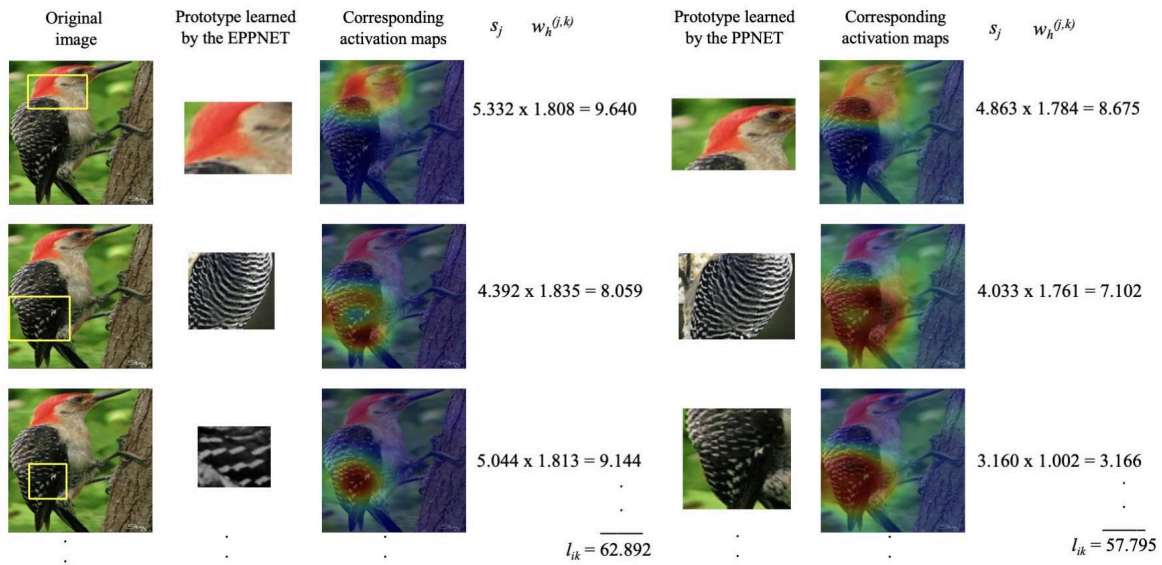

Fig. 4. Sample activation maps and their corresponding prototypes as learned by the proposed EPPNet (left) and the PPNet (right) from a training image.

Table 1. Accuracy (%) of the EPPNet and other xAI-based models for different backbones.

| Method | Backbone | | | | |
| --- | --- | --- | --- | --- | --- |
| | VGG19 | ResNet50 | ResNet152 | DenseNet121 | DenseNet161 |
| PPNet [18] | 73.23 | 77.18 | 78.42 | 75.40 | 78.11 |
| ProtoTree [24] | 68.70 | 78.56 | 71.20 | 73.20 | 72.40 |
| ProtoPool [21] | 78.40 | 84.10 | N/A | **81.50** | 81.00 |
| Deformable ProtoPNet [22] | 77.90 | **86.40** | 79.60 | 79.00 | 81.20 |
| EPPNet | **78.92** | 79.92 | **80.15** | 79.51 | **81.91** |

Comparison with the SOTA: We compare the EPPNet against several prototype-driven SOTA methods (in team of classification accuracy. The results of this experiment are tabulated in Table 1. It is evident that the EPPNet outperforms most competitors, particularly when VGG-19, ResNet152, and DenseNet161 are used as the backbone. For the DenseNet 121 backbone, the EPPNet achieves a very competitive performance. Fig. 4 illustrates sample activation maps and the corresponding prototypes as learned by the PPNet and EPPNet from a training image. Based on our experiments, we note that as we increase the number of squared $L2$ distances used by the mean-cluster loss in Eq. 4; i.e., as we increase $\theta$, the learned prototypes depict more specific regions within the images. On the other hand, the activation maps of PPNet tend to exhibit a higher degree of dispersion, as shown in Fig. 4. Fig. 4 also shows some of the similarity scores, $\{s_j\}$, and weights of the FC layer, $\left\{w_h^{(j,k)}\right\}$, used to compute the logit $l_{ik}$. Note that the logit value computed by the EPPNet is higher than that computed by the PPNet, which shows that for this training image, the EPPNet discovers prototypes that are more relevant.

Explainability after pruning: As highlighted in [17], pruning is important in terms of explainability. Pruning involves the random removal of several learned prototypes per class, after which a model is evaluated using the remaining learned prototypes. We prune half of the learned prototypes and evaluate the EPPNet and the PPNet. From Table 2, we can argue that the PPNet experiences a significant decrease in classification accuracy in the range of $[1, 5]\%$ after pruning. In contrast, for the EPPNet, the decrease in accuracy is comparatively minor and in the range of $[1, 2]\%$. This experiment confirms that the EPPNet learns meaningful prototypes that substantially contribute to the classification outcome.

Faithfulness score: We compute the faithfulness score for the ten classes that are the hardest to classify by the EPPNet and the PPNet. For each of these classes, we compute the faithfulness score by using 20 test images per class and the 10 learned prototypes per class. Table 3 demonstrates that EPPNet's faithfulness scores significantly surpass those of the PPNet for the majority of these classes, underscoring its superior explainability. Out of the ten classes, the EPP-

Table 2. Accuracy (%) of the PPNet and the EPPNet for different backbones after pruning half of the learned prototypes.

|  | VGG19 | | ResNet50 | | ResNet152 | | DenseNe | |
|---|---|---|---|---|---|---|---|---|
|  | PPNet | EPPNet | PPNet | EPPNet | PPNet | EPPNet | PPNet | E |
| Befor | 73.23 | 78.92 | 77.18 | 79.92 | 78.42 | 80.15 | 75.40 | |
| After pruning | 72.88 | 78.31 | 74.13 | 78.44 | 75.50 | 78.61 | 75.29 | |

Table 3. Faithfulness scores ($\uparrow$) of the PPNet and the EPPNet for different backbones and the ten hardest-to-classify classes.

|  | VGG19 | | ResNet50 | | ResN |
| --- | --- | --- | --- | --- | --- |
| Class name | PPNet | EPPNet | PPNet | EPPNet | PPNet |
| Black footed Albatross | 0.6592 | **0.7254** | -0.4933 | **−0.2959** | -0.1481 |
| Laysan Albatross | -0.8832 | **−0.7177** | 0.5885 | **0.6038** | 0.2054 |
| Sooty Albatross | -0.9053 | **−0.2089** | **−0.2161** | -0.3900 | -0.4785 |
| Groove billed Ani | 0.6832 | **0.9936** | **0.1727** | -0.0664 | 0.2794 |
| Pileated Woodpecker | **0.6697** | 0.5511 | 0.0944 | **0.8182** | -0.9277 |
| Indigo Bunting | 0.4919 | **0.5320** | 0.3771 | **0.4062** | **0.8973** |
| Yellow breasted Chat | **0.9538** | 0.8538 | 0.3010 | **0.4304** | 0.6814 |
| Shiny Cowbird | 0.3296 | **0.5510** | 0.5450 | **0.6090** | 0.7788 |
| Purple Finch | **0.6800** | 0.3611 | -0.8862 | **−0.8583** | 0.3468 |
| Heermann Gull | 0.2164 | **0.2794** | **0.3700** | 0.3052 | 0.4034 |

Table 4. Accuracy (%) of the EPPNet for different values of $\theta$ (mean-cluster loss) when 10 prototypes are learned per class.

|  | Backbone | | | | |
| --- | --- | --- | --- | --- | --- |
|  | VGG19 | ResNet50 | ResNet152 | DenseNet121 | DenseNet161 |
| $\theta = 1$ | 73.26 | 77.18 | 78.42 | 75.40 | 78.11 |
| $\theta = 3$ | 75.25 | 77.32 | 78.28 | 76.21 | 78.42 |
| $\theta = 4$ | 76.56 | 77.44 | 78.80 | 76.91 | 79.00 |
| $\theta = 5$ | 77.04 | 78.71 | 79.62 | 79.06 | 80.03 |
| $\theta = 10$ | **78.92** | **79.92** | **80.15** | **79.51** | **81.91** |
| $\theta = 20$ | **78.93** | **79.91** | **80.16** | **79.50** | **81.94** |
| $\theta = 30$ | **78.90** | **79.91** | **80.16** | **79.53** | **81.90** |

Net outperforms the PPNet in 7 of them when VGG-19 and ReseNet-50 are used as the backbone, and it performs at its best when ResNet-152, DenseNet-121 & DenseNet-161 are used as the backbone.

Ablation studies: We analyze how the EPPNet's accuracy varies with changes in $\theta$ in Eq. 4, where $\theta$ represents the number of squared distances used by our mean-cluster loss, $\mathcal{L}_{\text{mclst}}$. The results in Table 4 indicate that the EPPNet's best performance is attained when our mean cluster loss uses $\theta = 10$; which equals the number of prototypes learned per class. For this particular dataset, if the value of $\theta$ is higher than 10, the EPPNet tends to overfit and there is no visible improvement in accuracy. This is expected as only 10 distinct prototypes are learned per class. Note that in our mean-cluster loss, it is possible to associate a region $r$ multiple times with distinct prototypes if the resulting squared $L2$ distances are among the $\theta$ minimum distances. Based on this fact, when $\theta > 10$ in this study, the mean-cluster loss is very likely to use some of the regions multiple times but with distinct pro- totypes. As a consequence, the use of more squared $L2$ distances in $\mathcal{L}_{mclst}$ than prototypes learned per class does not provide further improvements. Also note that if $\theta = 1$, the EPPNet achieves the same accuracy as that achieved by the PPNet as our $\mathcal{L}_{mclst}$ loss becomes the loss $\mathcal{L}_{\text{clst}}$ used by the PPNet (see Eq. 7).

## 4. CONCLUSIONS

This work introduced the EPPNet, an active xAI-based method for image classification that leverages prototypes for enhanced explainability. The EPPNet uses a novel meancluster loss that relies on multiple prototypes to distinguish between classes more effectively. This particular loss not only boosts classification accuracy compared to other prototypical networks but also helps to learn prototypes that point to more specific image regions. This work also introduced the faithfulness score to measure, on a per-class basis, the relevance of prototypes in ensuring explainability, while accounting for the classification performance. Our experiments on the CUB-200-2011 dataset confirm the EPPNet's strong performance in terms of classification results and explainability. Our future work includes integrating sub-space extraction for objects within images, aiming at improving both accuracy and explainability.